\documentclass[letterpaper, 10 pt, conference]{ieeeconf}  %

\IEEEoverridecommandlockouts                              %

\usepackage{graphics} %
\usepackage{epsfig} %
\usepackage{times} %
\usepackage{amsmath} %
\usepackage{amssymb}  %
\usepackage{makecell}
\usepackage{multirow}
\usepackage{hyperref}

\usepackage{mathtools}
\usepackage{mathrsfs}
\usepackage{hyperref}
\usepackage{float}
\usepackage{pifont}%
\newcommand{\cmark}{\ding{51}}%
\newcommand{\xmark}{\ding{55}}%

\title{\LARGE \bf
Memorized action chunking with Transformers: Imitation learning for vision-based tissue surface scanning
}

\author{Bochen Yang$^{*\dagger1,2}$,        
        Kaizhong Deng$^{*1,2}$,
        Christopher J Peters$^{2}$,
        George Mylonas$^{1,2}$,
        Daniel S. Elson$^{1,2}$ %
\thanks{$^{*}$Equal Contribution}%
\thanks{$^{\dagger}$Work done during master study }%
\thanks{$^{1}$Hamlyn Centre for Robotic Surgery, Institute of Global Health Innovation, Imperial College London}%
\thanks{$^{2}$Department of Surgery and Cancer, Imperial College London, Exhibition Road, London, SW7 2AZ, UK. Corresponding email: {\tt\small daniel.elson@imperial.ac.uk}}%
}

\usepackage{eso-pic} 

\newcommand\AddNote{%
  \AddToShipoutPictureBG*{%
    \AtPageUpperLeft{%
      \parbox[t][\paperheight]{\paperwidth}{%
        \vspace*{1cm} 
        \hspace*{1cm} 
        \textbf{Note:} This work has been submitted to the IEEE for possible publication. Copyright may be transferred without notice, after which this version may no longer be accessible.
      }%
    }%
  }%
}

\begin{document}

\AddNote

\maketitle

\thispagestyle{empty}
\pagestyle{empty}

\begin{abstract}

Optical sensing technologies are emerging technologies used in cancer surgeries to ensure the complete removal of cancerous tissue. While point-wise assessment has many potential applications, incorporating automated large-area scanning would enable holistic tissue sampling. 
However, such scanning tasks are challenging due to their long-horizon dependency and the requirement for fine-grained motion.
To address these issues, we introduce Memorized Action Chunking with Transformers (MACT), an intuitive yet efficient imitation learning method for tissue surface scanning tasks. It utilises a sequence of past images as historical information to predict near-future action sequences. In addition, hybrid temporal-spatial positional embeddings were employed to facilitate learning. 
In various simulation settings, MACT demonstrated significant improvements in contour scanning and area scanning over the baseline model. In real-world testing, with only 50 demonstration trajectories, MACT surpassed the baseline model by achieving a $60\!-\!80\%$ success rate on all scanning tasks.
Our findings suggest that MACT is a promising model for adaptive scanning in surgical settings.
\end{abstract}

\section{INTRODUCTION}

 Completely resecting the tumour while preserving maximal healthy tissue is one of the most critical objectives for surgery. This could lead to minimal local recurrence and prolong disease-free survival. To ensure optimal results, it is necessary to establish a negative circumferential resection margin to guarantee that no tumour is present at the edges of the removed specimen. Thus, there is a pressing demand for \textit{in vivo} tissue assessment tools capable of characterising tissue in real time.
 
 Recently, several optical sensing technologies have emerged to facilitate real-time, non-invasive tissue classification, including Diffuse Reflectance Spectroscopy (DRS)~\cite{Nazarian2022JAMA,Nazarian2024IJS}, Raman spectroscopy~\cite{raman3}, and confocal endomicroscopy~\cite{pcle_review}. 
 These techniques rely on contact-based sensing instruments which require positioning of the sensing tip at the interested point to assess the tissue. 
They are usually designed as hand-held probes, allowing manual positioning at the target tissue by the operating surgeon. Their acquisition protocol usually requires steady contact to obtain high quality data. 
 
 Despite their success on single-site sampling, their utility on large-area sampling \textit{in vivo} has not been realised. For instance, manual scanning of the target area with the DRS probe may lead to high variance in the analysed results because of the non-ergonomic control, unsteady movement, and variable contact pressure~\cite{DRS_pressure}. An automated robot-assisted scanning system could be beneficial in extrapolating to large-area tissue scanning scenarios. Specialised control algorithms and setups may achieve this, but they require extensive hand-crafted engineering and constrained working conditions~\cite{scan_rotation_rob_raman}.

Imitation learning is capable to learn skills with minimal prior knowledge in achieving complex surgical scanning tasks. 
It can be used for learning the microscopic probe scanning preferences of experienced surgeons~\cite{scan_in_pCLE_LfD}, learning ultrasound scanning for carotid artery examination from demonstrations~\cite{USImitation}, and learning drop-in gamma probe scanning in laparoscopic surgery~\cite{gamma_probe}.
However, these methods cannot be applied to this probe scanning task because of the format of the instrument readout and contact requirement.
A more sophisticated neural network with a better state perception is a promising approach to solve this problem.

Recent transformer-based models have demonstrated their capability to model long sequences. %
Action Chunking with Transformers (ACT) is an advanced Transformer-based architecture which integrates temporal ensemble mechanisms and employs a conditional VAE (CVAE) to model human demonstrations~\cite{ACT}. It outperforms previous imitation learning algorithms in learning from a limited number of human demonstrations in daily life tasks. Despite its potential in learning complex tasks, it does not generalise well to this tissue scanning task as it lacks awareness of the overall trajectory. %

This study introduces a framework for robot-assisted autonomous optical instrument scanning for \textit{in vivo} abdominal tissue. We chose an exemplar application of phantom liver scanning, although the findings could be applied to other surgical scenarios. The probe model that was scanned across the surface could be substituted by another probe-based technique. This work has made the following main contributions: 
\begin{itemize}
    \item The Memorized Action Chunking with Transformer (MACT) based on ACT~\cite{ACT} is proposed, which introduces a memory mechanism to utilise historical colour and depth images from the robot wrist camera to predict action sequence.
    \item A hybrid temporal-spatial Positional Embedding is introduced to explicitly model the interrelations among transformer input tokens.
    \item A robotic-assisted optical instrument scanning platform was built for human demonstration collection and autonomous scanning evaluation.
    \item MACT was evaluated on both simulation and real-world setups and shown to outperform the baseline ACT among all tested tasks.
\end{itemize}

\begin{figure*}[!t]
\vspace{5pt}
\includegraphics[width=\linewidth]{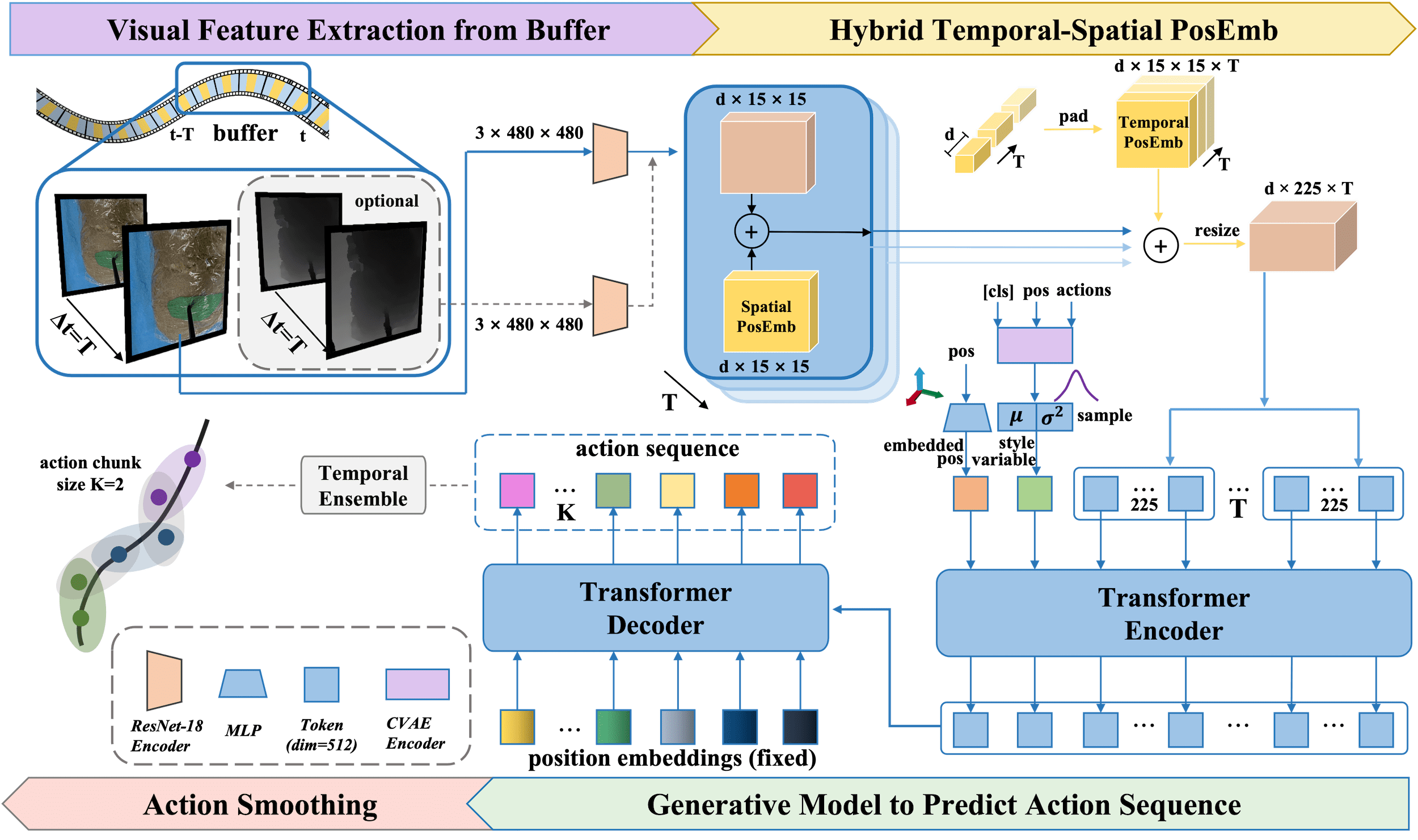}
\vspace{-20pt}
\caption{\textbf{Architecture of Memorized Action Chunk with Transformers (MACT).} MACT buffered the past $T$ RGB and depth images from a mount-on camera for a ResNet-18 encoder to extract high-dimensional visual features. A Spatial PosEmb which was identical along the time horizon axis and Temporal PosEmb which was identical on the spatial axis were added to visual features to form image tokens. The transformer encoder took image tokens, current position, and CVAE style variables to produce encoded tokens to feed into a transformer decoder to predict the next $K$ action sequences. Temporal ensemble and action chunking further smoothed the predicted action sequence} 
\label{FIG_1}
\vspace{-15pt}
\end{figure*}

\section{PROBLEM STATEMENT}
\textbf{Task Description.} Tissue scanning can be achieved from two perspectives: area scanning and contour scanning. Area scanning involves sampling all points within an enclosed boundary by moving the probe along a grid-like coverage path. Alternatively, contour scanning aims to track the boundary of a specified area of interest to sample the border status. Area scanning can identify the area of suspicious tissue, while contour scanning can assess whether a proposed resection margin is free of suspicious tissue. 
Both methods could contribute to ensuring a negative resection margin following tumour removal. The area of interest can be manually indicated by the surgeon or derived from an external imaging system capable of detecting the suspicious area, such as near-infrared fluorescence imaging~\cite{flo}. 

\textbf{Problem Statement.} The aim was to build a policy which imitated human demonstrations of tissue scanning tasks. The policy was denoted as $\pi_{\theta}(a_{t} | s_{t})$, where the observed state $s \in \mathbf{S}$ and generated action $a \in \mathbf{A}$ was for a single step. The policy could include a sequence of historical observations $s_{t-T:t}$ and predict a sequence of future actions $a_{t:t+K}$, thus denoted as $\pi_{\theta}(a_{t:t+K} | s_{t-T:t})$,  where $K$ represents the length of the predicted action sequence, and $T$ denotes the length of the input image sequence. The trajectory ${\tau}$ generated from $\pi$ passed a series of waypoints, ${\{ \prescript{\pi_\theta}{}{p_{i}} \}}^M_{i=1}$. Specifically, in the scanning task, the target area $\mathcal{A}$ was uniformly sampled to generate discrete points $\{ q_j \}^N_{j=1} \in \mathcal{A}$ across either surface or its boundary. The objective was defined as finding an optimal policy which can cover more points in the target area, thus represented as:
\begin{equation} \label{eq:objective}
\max_{\pi_\theta} \sum_{\substack{\mathcal{A}}} \sum_{i=1}^{M} \sum_{j=1}^{N} \frac{\delta(\| \mathbf{p}_i - \mathbf{q}_j \| \leq d)} {N} \times 100 - \mathcal{L}
\end{equation}
where $\delta(condition)$ returned 1 if conditions were satisfied, $d$ was the acceptance distance for a successful reaching, and the $\mathcal{L}$ represented the distance penalty. In the proposed framework, imitation learning was used to train a policy $\pi_{\theta}$ with collected human expert trajectories dataset ${\tau^*}\in D^*$, hence optimising its performance.

\section{METHOD}
This section describes the construction of the MACT framework by introducing its main components: the temporal ensemble in \autoref{subsec:te}, the memory-mechanism in \autoref{subsec:memory}, hybrid temporal-spatial positional embeddings in \autoref{subsec:hpe} and other details of network architecture in \autoref{subsec:MACT}.

\subsection{Temporal Ensemble}\label{subsec:te}
A temporal ensemble approach was proposed to mitigate compounding errors \cite{ACT}. 
For a sequence of $N$ future action $\{ a_i | i=0,...,N-1 \}$, 
the temporal ensemble is a weighted sum of actions defined as: 
$A_{\text{agg}} = \sum_{i=0}^{N-1} w_i \cdot a_i$.
These weights are calculated as $w_i = \frac{\exp(-\alpha \cdot i)}{\sum_{j=0}^{N-1} \exp(-\alpha \cdot j)}$, indexed future actions. $\alpha$ is a constant that determines the rate of exponential decay. Despite its effectiveness for many real-world manipulation tasks, it still faces challenges with certain tasks, as discussed in \autoref{discussion}.

\subsection{T-step Memory-enhanced ACT}\label{subsec:memory}

MACT aims to generate decisions informed by both past context and the future objective. This was achieved by predicting a sequence of actions over a future window $K$ and analyzing the historical context of states $s_{t-T:t}$, which extended back $T$ time steps from the current state. MACT policy $\pi_{\theta}(a_{t:t+K} | s_{t-T:t})$ is distinct from ACT policy $\pi_{\theta}(a_{t:t+K} | s_{t})$~\cite{ACT}, open-loop diffusion policy $\pi_{\theta}(a_{t:t+K} | s_{t})$, 
or closed-loop policy with $T$-frame stacking $\pi_{\theta}(a_{t} | s_{t-T:t})$. %
It integrates a broader temporal perspective which can enhance the decision-making process.

In practice, a buffer was created to store camera images from the last $T$ steps during both the training and inference phases. 
If a historical image is missing, the earliest available image within the buffer is used as a form of padding to ensure input sequence integrity.

\subsection{Hybrid Temporal and Spatial Positional Embeddings}\label{subsec:hpe}
Transformer architectures are permutation-invariant and require additional information to represent the relative positional information in a token sequence. 
Temporal encoding injects the time sequence dependencies for visual observations while spatial encoding enhances the spatial information of feature maps.
In sequences of consecutively captured images within a short timeframe, extracted feature maps often exhibit high similarity, which makes it challenging to model their dependencies.
To address this challenge, a mechanism enabling the model to discern temporal sequences and frame order was introduced via the integration of sinusoidal positional encoding embedded into image features, which was essential for the accurate analysis of tissue and end-effector movements. 
This approach not only assisted the transformer encoder in identifying the temporal sequences of tokens for each frame but also facilitated the convergence of the transformer model. 

Spatial positional embeddings assign a unique vector with a dimensionality of $d$ to every image spatial dimension across the height and width of the feature map $(d{\times}15{\times}15 )$ as described in~\cite{DETR}.
The Temporal Positional Embedding matrix~\cite{embeddings} \(\mathcal{T}_{T \times d}\) was defined in ~\autoref{TPE} where \(T\) represents the input of the furthest past image and $d$ is set to 512, with $t$ denoting the time from 0 to $T$ and $i$ being the dimension ranging $0$ to $d$.
\begin{equation} \label{TPE}
\textstyle \mathcal{T}_{(t, 2 i)}=\sin \left(\frac{t}{10000^{2 i / d}}\right) \! \quad \mathcal{T}_{(t, 2 i+1)}=\cos \left(\frac{t}{10000^{2 i / d}}\right)
\end{equation}

To integrate these two types of Positional Embeddings (PosEmb), as illustrated in \autoref{FIG_1} (top right corner), the process initiated by encoding the width and height of the current feature map with $d \! \times \! H \! \times \! W$ spatial embeddings, which were then added to the feature map. Next, the temporal PosEmb for the corresponding time step was generated and reshaped to match the shape of spatial PosEmb before addition. %
This process was repeated $T$ times, resulting in an output of $d \! \times 15\! \times 15\! \times T$. Finally, the dimensions for width and height were flattened to produce a shape of $d \! \times 225\! \times T$. A matrix additive was adopted similarly to \cite{transformer}, utilising the additive nature of sine and cosine functions. This technique helped the learned position embeddings at the input of the transformer decoder to recover encoded temporal-spatial relationships, thereby enhancing the model's ability to interpret and utilise the integrated positional information effectively.

\subsection{Network Architecture: MACT with CVAE}
\label{subsec:MACT}

Previous research has demonstrated the importance of training with CVAE to model the high-dimensional variation in human operations~\cite{ACT}. The same implementation was adopted here. 
The input to the transformer encoder consisted of multiple tokens with a dimensionality of 512. These included the processed visual tokens of the past $T$ images, along with the current Cartesian position and the style variable. %
The transformer decoder employed cross-attention on the output of the encoder to predict action token sequence in $K\! \times 512$. These tokens were then mapped as target Cartesian positions sequence for the next $K$ steps. Additionally, a temporal ensemble module further smoothed the action sequences.%

MACT used a single ResNet18 backbone sequentially tokenising the past $T$ images. It could be further extend to incorporate depth images or inputs from additional cameras. These images were resized and duplicated on its channel to match the input format for adapting to another ResNet18 visual tokenizer.
Despite the diversity of camera inputs, a uniform spatial-temporal PosEmb was applied across backbones to ensure consistent temporal and spatial information preservation without differentiating camera views.

\section{EXPERIMENTS}
\subsection{Simulation environment and path generator}
Pybullet served as the physics simulator for executing robot movements and gathering state information, including RGB images, depth images, positions, and tool-tissue penetration distances. Five tissue models with unique shapes and textures were reconstructed from stereo images to validate the generalization capacity. 

To create an example 3D target area surface on the tissue, a 2D irregular region (random shape and size) was initially defined on an image from the wrist camera, simulating a user-created suspicious region of interest, or potentially automatically generated from a visual information or near-infrared fluorescence image as described in \autoref{subsec:real}. 3D positions and normal vectors were obtained by ray-casting the sampled points from the 2D region onto the tissue. Subsequently, the surface was reconstructed through Delaunay triangulation and dynamically imported into the simulation environment upon each reset. Additionally, six image augmentations were applied, including adjustments in brightness, shearing, blurring, colour shifting, random noise, and regional dropout. %

In the simulation, three tasks were assessed: single tissue contour scanning, single tissue raster scanning, and multiple tissue raster scanning. 50 demonstration trajectories were collected for each task.
Contour scanning involved the robot tracing the boundary of the targeted area, examined on a single tissue model. The waypoints of target trajectory were mapped from projected points of the contour. %
Raster scanning entailed the robot performing a raster scan across the area of interest. It was evaluated on both single and all five tissue models to assess its generalizability. The raster scanning was achieved by moving the end-effector from the upper to lower edge of the target surface in parallel lines, slightly offset and bounded by the extracted 3D contours to account for the topography of the surface.%

\subsection{Real-world setting and data collection} \label{subsec:real}
\begin{figure}[htbp]
    \centering
    \vspace{-5pt}
    \includegraphics[width=\linewidth]{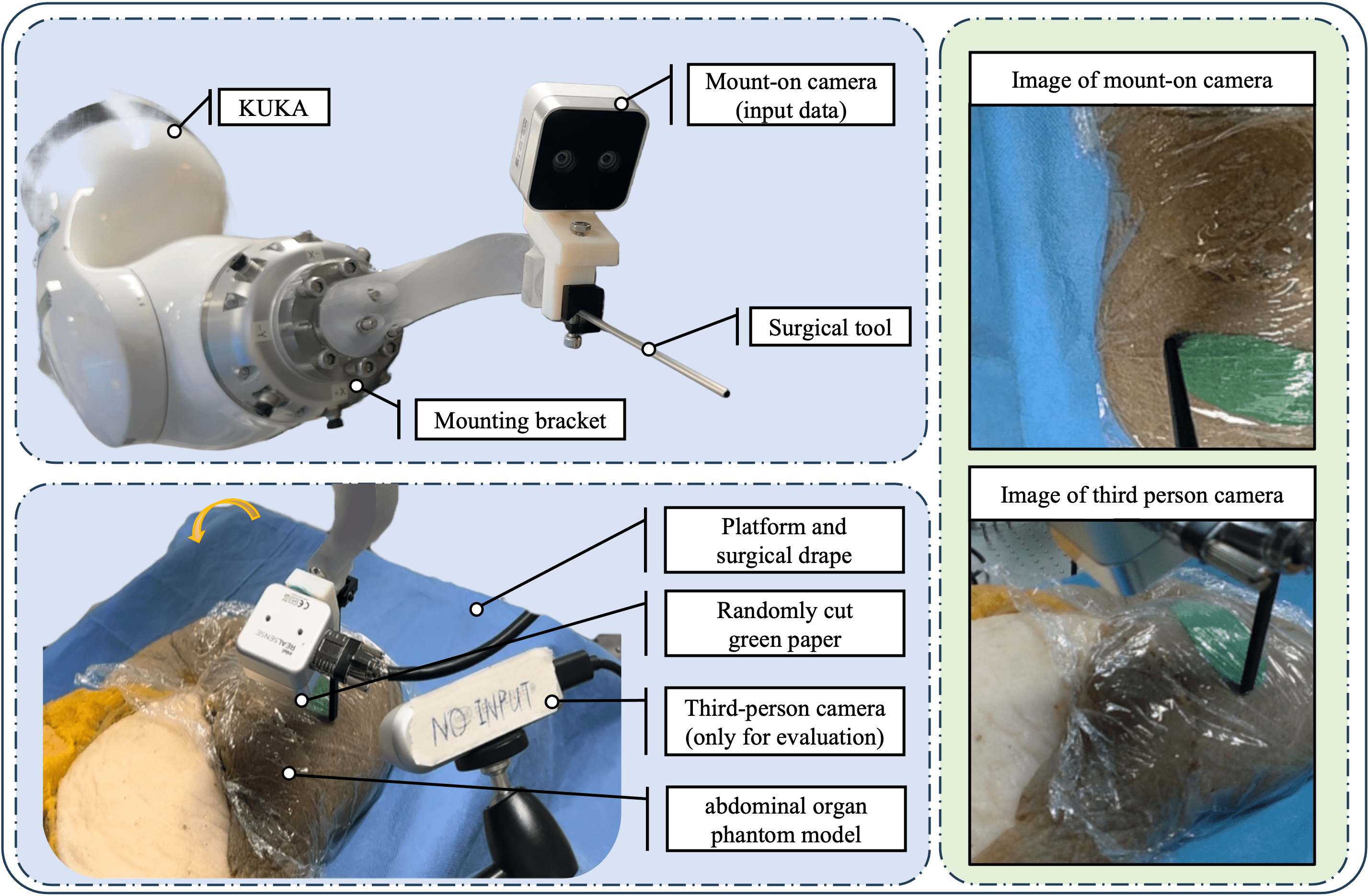}
    \vspace{-20pt}
    \caption{Real-world scanning task environment with randomly-cut green paper on an abdominal organ phantom. The wrist camera was mounted onto the KUKA with a 3D-printed bracket with a static third-person camera for monitoring. }
    \label{FIG_2}
    \vspace{-10pt} %
\end{figure}
The real-world setting is shown in \autoref{FIG_2}. This research utilised a KUKA LBR iiwa14 R820 robot arm. Its motion control and ROS interface were supported by \textit{iiwa\_stack}~\cite{iiwa_stack}. The end-effector incorporated a probe model and a wrist camera, with a mounting bracket to attach to the media flange of the KUKA robot. The probe model was a 3D-printed black cylinder ($0.5 \, cm$ diameter, $10 \, cm$ length) approximating the geometry of a clinically approved DRS probe~\cite{Nazarian2022JAMA} but with a relatively soft tip ($2 \, cm$) for safety reasons. An Intel RealSense D405 camera was configured as a wrist camera mounted on the end-effector to capture RGB and depth information within a $7-30\,cm$ range, capturing images with dimensions of $480\times \! 480 \times \! 3$ (RGB) and $480 \times \! 480$ (depth) after cropping. The mounting bracket's S-shaped design was ergonomically crafted to facilitate manual guidance of the robot arm.

A silicone abdominal phantom model was placed on blue surgical drapes. A translatable platform was placed under the drapes to allow motion of the phantom. Target scanning surfaces were simulated using 10 uniquely cut pieces of green paper, representing either a visual target for scanning. %
These were randomly overlaid on the tissue each time. %
Plastic 'cling film' was used to secure the paper on the surface, minimising displacement during interactions.

\begin{figure*}[t]\centering
\vspace{5pt}	
 \includegraphics[width=\linewidth]{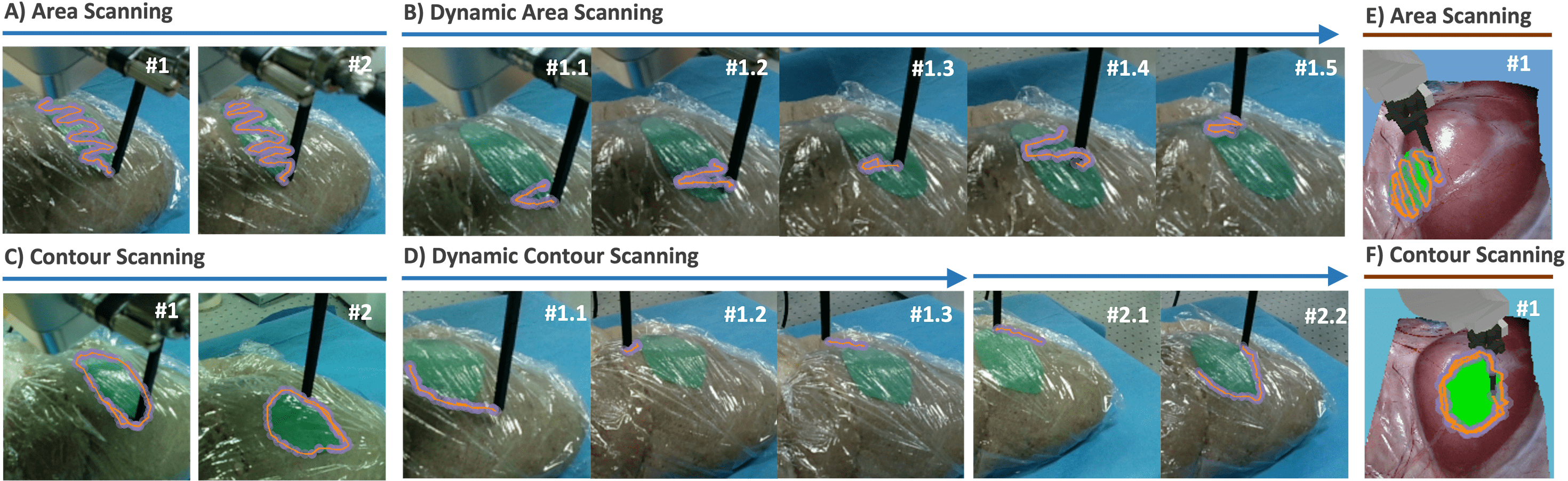}
    \vspace{-20pt}
	\caption{Real-world experimental results of MACT for different tasks (left). The phantom model was fixed in tasks A \& C, while it included a sudden planar translation during tasks B \& D. For static tasks, each frame represents a single location for probe surface sampling. For dynamic tasks, each frame represents a section trajectory after a planner shift of the platform. Simulated experimental results of MACT for tasks E \& F (right). \textbf{\#} stands for different cases and the horizontal axis represents the timeline}
    \label{FIG_3}
    \vspace{-15pt} %
\end{figure*}

The human demonstrations were recorded with manual robot guidance. The KUKA robot was configured with a fixed end-effector orientation, allowing transitional movement only. After 30 minutes of training, operators dragged the S-shaped bracket, maintaining contact of the end effector with the 3D surface. Data collection involved randomizing the initial 3D position of the end-effector and the phantom's planar positions to ensure variability.

The study investigated both contour and raster scanning tasks under static and dynamic conditions, resulting in four distinct tasks. Initially, 30 trajectories were collected for each task with a fixed platform position. To simulate dynamic environments, the planar position of tissue was altered by suddenly rotating the platform beneath it. For this setup, 20 new trajectories were collected, combined with 30 trajectories from the fixed environment, yielding a total of 50 for each dynamic task. This aimed to disrupt the current raster scanning process, prompting the MACT model to adjust to the dynamic environment. The collection frequency was set at $4 \, Hz$, with each contour and raster scanning trajectory lasting for 100 and 150 steps correspondingly under both static and dynamic scenarios.

\subsection{Training setup}

The training of MACT utilised similar paradigm as ACT with L1 reconstruction loss and ten times of KL divergence loss. ACT has been implemented as the baseline model to compare. It can be viewed as a special case for MACT with $T\!=\!1$. For simulations, demonstration trajectories were modulated to a $5 \, mm$ step size. The camera resolution was set to $240 \! \times \! 240$ with $T\! =\! 30$ to include past steps. The training was carried out on RTX 4090 (24G) with a batch size of 6 for 3000 epochs. For real-world tasks, demonstration trajectories were sampled at 4 FPS with $480 \! \times \! 480$ resolution images. It was trained for 5000 epochs with batch size of 2. 

\subsection{Testing metrics}
In simulations, we evaluated scanning performance using a metric that combines the surface coverage ratio and distance penalty. It is expressed as \autoref{eq:objective}. The first term is the ratio of points scanned within a $d=5\,mm$ working radius to all sampled points. The second term is derived as $\mathcal{L}\!=\!0.5\!\times\!e$ where $e$ is the penetrating or deviating distance error measured in millimetres. 

In the real-world testing, the success rate was adopted to assess model performance. Success was defined as maintaining proper contact conditions and following the expected scanning trajectory without significant deviation. Task success hinged on three qualitative criteria: the alignment of the sweeping movement amplitude with the area contour, the coverage rate of this area, and avoiding excessive compression of the tissue or loss of contact with the scanned surface ($<1\,cm$). The success rate can effectively represent the overall performance of each model, as there is often a distinguishable difference in their behaviour. 

\section{RESULTS}

Scanning performance for simulated tasks is shown in \autoref{table:sim}. An example of scanning in simulation is shown in \autoref{FIG_3}.
A scripted expert leveraging state information and geometry of the surface model in the simulator worked as a reference for the best available agent. 
Compared to the baseline model ACT, the proposed MACT with the memory mechanism outperformed the baseline in all settings among all tasks. %
For the contour scanning task, performance was not sensitive to $T$ but it relied on the presence of depth information, which could improve it from $71.5$ to $85.9$ with only $10\%$ below scripted expert.
For the raster scanning task, it is notable that the single tissue scanning required more past information, $T\! =\! 30$, for optimal performance. With both the historical information and depth images, the generalisation capability of scanning on 5 tissues could be improved up to $66.3$. To balance the performance and computation cost, MACT$^\star$ with $T\!=\!15$ and $K\!=\!5$ was used as the standard configuration to test in the real world tasks.

\begin{table}[h]
\vspace{-5pt}
\centering
\caption{Scanning performance for simulated tasks}
\label{table:sim}
\vspace{-5pt}
\resizebox{\columnwidth}{!}{%
\begin{tabular}{l|cccc|c|cc}
\hline
\multirow{2}{*}{Methods} & \multirow{2}{*}{$T$} & \multirow{2}{*}{$K$} & \multirow{2}{*}{Depth} & \multirow{2}{*}{TE} & \multicolumn{1}{c|}{Contour} & \multicolumn{2}{c}{Raster} \\ 
 & & & & & 1 tissue & 1 tissue & 5 tissues \\ \hline
Script & - & - & - & - & 95.9 & 88.1 & 85.2 \\ \hline
ACT & 1 & 5 & \xmark & \xmark & 53.1 & 28.1 & 15.6 \\
MACT & 5 & 5 & \xmark & \xmark & 71.2 & 33.1 & 25.9 \\
MACT$^\star$ & 15 & 5 & \xmark & \xmark & 71.5 & 46.8 & 58.5 \\
MACT & 30 & 5 & \xmark & \xmark & 69.7 & \textbf{62.9} & 52.3 \\
MACT & 15 & 5 & \cmark & \xmark & \textbf{85.9}  & 55.6 & 60.5 \\
MACT & 15 & 5 & \cmark & \cmark & 72.6 & 41.7 & \textbf{66.3} \\
MACT & 15 & 30 & \cmark & \cmark & 80.6 & 56.4 & 57.6 \\
 \hline
\end{tabular}%
}
\vspace{-5pt} 
\end{table}

Success rates for different real-world tasks are shown in \autoref{table:real-world-task}. \autoref{FIG_3} shows some successful examples of MACT performing scanning tasks, with their tool-tip trajectories plotted in the image view.
With $K\! =\! 5$, ACT performs poorly across all tasks, especially in raster scanning tasks. Extending the prediction to $K \! = \! 20$ leads to improved ACT performance, attributed to the network's capacity to imitate longer action sequences that capture the pattern of raster scanning during training. In contrast, our proposed MACT method outperforms baselines on all tasks. Besides, the integration of a depth camera and predicting short action sequences could be beneficial in dynamic scanning tasks. This was because the sudden shifts or rotations could prohibit the 3D perception from history and the agent need to rely on instant geometrical information to recover from the disturbance.

\begin{table}[h]
\vspace{-5pt}
\centering
\caption{Success rate for real-world tasks}
\label{table:real-world-task}
\vspace{-5pt}
\resizebox{\columnwidth}{!}{\large
\begin{tabular}{l|cc|c|cc|cc}
\hline
\multirow{2}{*}{Methods} & \multirow{2}{*}{$T$} & \multirow{2}{*}{$K$} & \multirow{2}{*}{Depth} & \multicolumn{2}{c|}{Contour} & \multicolumn{2}{c}{Raster} \\  
 &  &  &  & Static & Dynamic & Static & Dynamic \\ \hline
ACT & 1 & 5 & \xmark & 2/5 & 0/5 & 0/10& 0/10 \\
ACT & 1 & 20 & \xmark & 3/5 & 1/5 & 5/10 & 1/10 \\
MACT$^\star$ & 15 & 5 & \xmark & 3/5 & 2/5 & \textbf{8/10} & 4/10 \\
MACT & 15 & 5 & \cmark & \textbf{4/5} & \textbf{3/5} & \textbf{8/10} & \textbf{6/10} \\ \hline
\end{tabular}%
}
\vspace{-10pt}
\end{table}

Some previous trials struggled with raster scanning, often causing unintended tissue deformation and straying from the designated green area early on. For instance, as illustrated in \autoref{FIG_5}(a) and (b), two common failure cases were observed when employing the ACT with $T\!=\!1$ and $K\!=\!5$. 
The first case tended to move in the right direction toward the end but without the necessary sweeping to cover the area. The second case failed to approach the contour at the beginning resulting in an oscillating motion near the initial point. The MACT can successfully avoid these two scenarios.

The effectiveness of the hybrid temporal-spatial PosEmb was validated through the following ablation study using MACT$^\star$. During training, the MACT$^\star$, which included hybrid temporal-spatial PosEmb, achieved a reduction in training loss from 0.27 to 0.16, compared to the MACT without it. It improved performance across all three simulation tasks by 1.4\%, 13.6\%, and 9.1\%. In the real world testing, although there was no significant improvement in contour scanning, it increased the raster scanning success rate from 70\% and 30\% to 80\% and 40\%.

\section{DISCUSSION}\label{discussion}

\begin{figure}[h]
    \centering
    \includegraphics[width=0.8\linewidth]{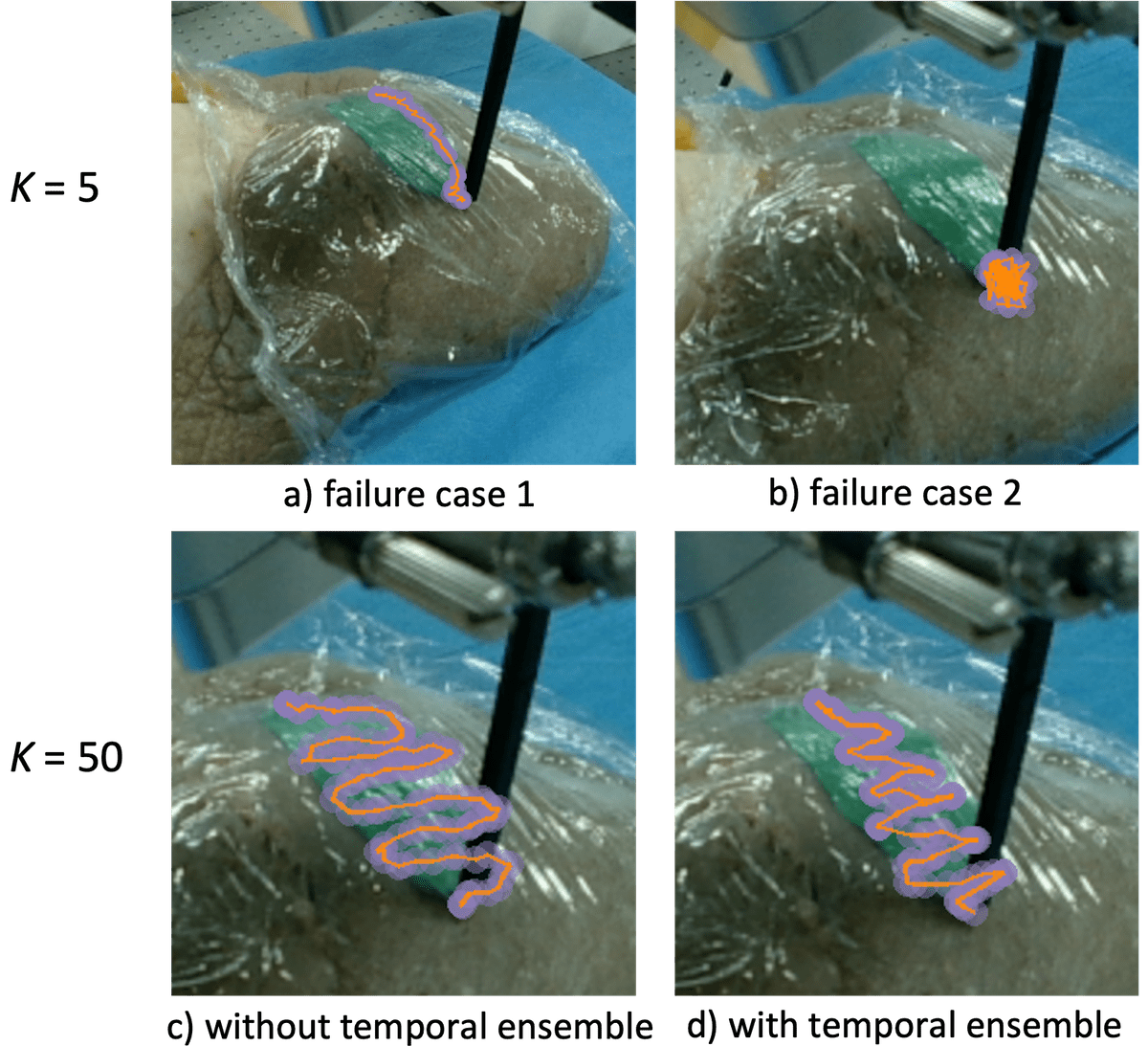}
    \vspace{-10pt} 
    \caption{Common ACT failure cases in a raster scanning with \(T=1\), \(K=5\) in a) and b). Examples ACT performance of \textit{w/o} and \textit{w/} temporal ensemble with \(K=50\) in c) and d).}
    \label{FIG_5}
    \vspace{-10pt} %
\end{figure}

\textbf{Temporal Ensemble.} In the experiments, it was discerned that temporal ensemble could inadvertently lead to diminished performance outcomes in scanning tasks. As illustrated in \autoref{FIG_5} (B), applying temporal ensemble could smooth out the necessary lateral sweeping when $K\!=\!50$. However, predicting long sequence is not always feasible, which complicates the application of ACT in surface scanning tasks.

\textbf{Effectiveness of $T$ and $K$.} Balancing the parameters of past steps $T$ and future steps $K$ of MACT is crucial. 
From our results, the historical information is crucial in tissue surface scanning tasks. A moderate value of $T\!=\!15$ yields a balance between its performance and training cost. 
The impact of $K$ is similar as described in \cite{ACT} in general. However, MACT could utilise history to lead to robust and smooth motion. This provides a chance to use smaller $K$ to closely align with closed-loop control, which shows robustness in a dynamic environment.
However, the setup could be adjusted to optimise its performance depending on the characteristics of diverse tasks when MACT is used in real clinical environments.

\textbf{Future Work.} The system could be tested with a more clinically realistic approach to identify the region of interest by using intrinsic or extrinsic fluorescence or other imaging systems \cite{flo} to replace the green marked in the experiment.
Despite the advantage of this generalised method, the implementation of classical specialised methods could be helpful to validate its performance. The use of a real optical probe in further evaluation will allow for the investigation of its clinical value.

\section{CONCLUSIONS}
In this work, we proposed a transformer-based imitation learning method, MACT, for robot-assisted tissue scanning for cancer margin identification. MACT could understand end-effector movement and environmental changes by leveraging historical RGBD visual information to predict a short action sequence. A hybrid temporal-spatial PosEmb integrated into MACT could improve its understanding of historical information. Through experiments in simulation and real-world settings, MACT successfully achieved autonomous contour and raster scanning tasks on static or dynamic tissues. %
Although the experiment was on a liver phantom, it could potentially be applied to different types of tissues and extrapolated to other probe-based measurement systems.

\section*{ACKNOWLEDGMENTS}
This paper is independent research funded by the National Institute for Health Research (NIHR) Imperial Biomedical Research Centre (BRC), the Cancer Research UK (CRUK) Imperial Centre and the Wellcome Trust ITPA MedTechOne awards. 
\vspace{-0.8em}
\bibliographystyle{IEEEtran}
\bibliography{IEEEabrv,root}

\end{document}